\documentclass{article}
\usepackage{spconf,amsmath,graphicx}
\usepackage{xcolor, bm, multirow}
\usepackage{amssymb}
\usepackage{subcaption}
\usepackage{xcolor}
\usepackage[colorlinks=true, allcolors=blue]{hyperref}
\usepackage{orcidlink}
\usepackage{etoolbox}

\title{Hyperspectral Image Classification using Spectral$-$Spatial Mixer Network}

 \name{Mohammed Q. Alkhatib $^{\orcidlink{0000-0003-4812-614X}}$ 
 }

\address{College of Engineering and IT, University of Dubai, Dubai, 14143, UAE \\mqalkhatib@ieee.org}

\begin{document}

\maketitle

\begin{abstract}
This paper introduces SS-MixNet, a lightweight and effective deep learning model for hyperspectral image (HSI) classification. The architecture integrates 3D convolutional layers for local spectral-spatial feature extraction with two parallel MLP-style mixer blocks that capture long-range dependencies in spectral and spatial dimensions. A depthwise convolution-based attention mechanism is employed to enhance discriminative capability with minimal computational overhead. The model is evaluated on the QUH-Tangdaowan and QUH-Qingyun datasets using only 1\% of labeled data for training and validation. SS-MixNet achieves the highest performance among compared methods, including 2D-CNN, 3D-CNN, IP-SWIN, SimPoolFormer, and HybridKAN, reaching 95.68\% and 93.86\% overall accuracy on the Tangdaowan and Qingyun datasets, respectively. The results, supported by quantitative metrics and classification maps, confirm the model’s effectiveness in delivering accurate and robust predictions with limited supervision. The code will be made publicly available at \url{ https://github.com/mqalkhatib/SS-MixNet}

\end{abstract}
\section{Introduction}
\vspace{-0.75em}
Hyperspectral imaging (HSI), available since the 1980s~\cite{qu2022review}, provides rich spectral and spatial information across hundreds of narrow, contiguous bands ranging from the visible to infrared spectrum. This wealth of data enables detailed remote sensing tasks that were previously unattainable. Consequently, HSI classification has become a vital area of research in Remote Sensing (RS), with broad applications in Earth Observation (EO), including land cover/use mapping and environmental monitoring. Achieving accurate and reliable classification requires effective extraction of both spectral and spatial features.

Recent advances in Deep Learning (DL) have driven significant progress in HSI classification, with Deep Convolutional Neural Networks (DCNNs) outperforming traditional techniques~\cite{alkhatib2023tri, roy2019hybridsn}. Early approaches, such as the 1D-CNN in~\cite{hu2015deep}, focused on extracting spectral features but overlooked spatial context. To address this, 2D-CNNs~\cite{makantasis2015deep} were introduced to incorporate spatial information. However, these methods could not fully exploit the three-dimensional nature of HSI data. This limitation was addressed in~\cite{hamida20183}, where 3D-CNNs were proposed to jointly capture spectral and spatial features. Building on this, HybridSN~\cite{roy2019hybridsn} combined 3D-CNN and 2D-CNN layers to enhance multi-level feature extraction. Similar hybrid strategies have since been explored~\cite{alkhatib2023tri, yu2020simplified}.

Inspired by the success of transformer architectures in natural language processing, researchers have extended their application to computer vision and EO~\cite{cai2022t}, including HSI analysis~\cite{roy2023multimodal}. While transformers offer strong modeling capabilities, they typically require large amounts of labeled data, which limits their effectiveness in remote sensing scenarios. To address this, HSIFormer~\cite{alkhatib2024hsiformer} introduced a Vision Transformer (ViT) with local window attention (LWA), improving accuracy under data-scarce conditions. Nonetheless, transformer models still incur significantly higher computational costs and demand more hardware resources than conventional CNN-based approaches.

\begin{figure*}[!t]
\centering
\includegraphics[clip=true, trim = 20 10 20 5,width= 0.85\linewidth]{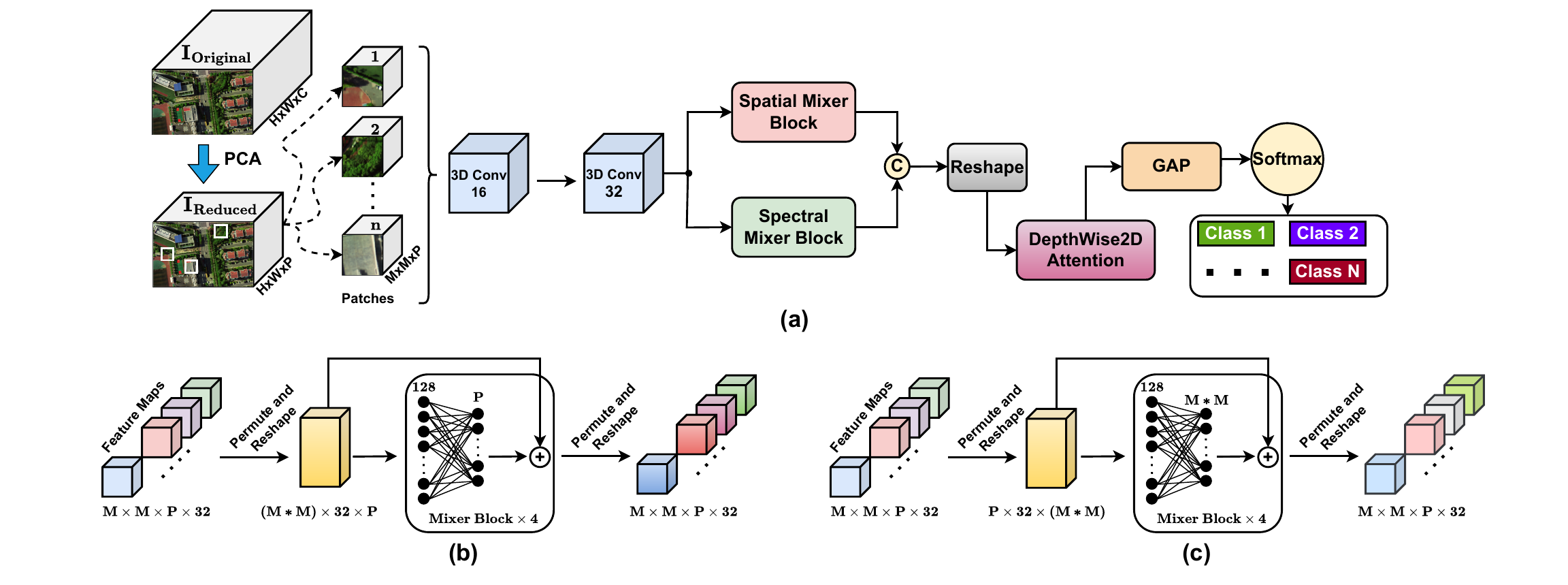}
\vspace{-1em}
\caption{(a) Overall architecture of the proposed  SS-MixNet Model; (b) Spectral Mixer Block; (c) Spatial Mixer Block.}
\vspace{-1em}
 \label{fig:model}
\end{figure*}

In this paper, a novel and lightweight framework, SS-MixNet, is proposed for hyperspectral image classification. The model operates on hyperspectral patches while maintaining spatial resolution and effectively decoupling spectral and spatial mixing through two parallel MLP-based modules. SS-MixNet is designed to be computationally efficient, with fewer parameters compared to conventional models. The architecture is inspired by MLPMixer~\cite{tolstikhin2021mlp}, integrating spectral and spatial mixers for long-range dependency modeling. Furthermore, a depthwise attention module is incorporated to enhance feature representation by adaptively reweighting spatial features on a per-channel basis.


\section{NETWORK ARCHITECTURE} \label{sec:model}
\vspace{-0.75em}
The architecture of the proposed model, shown in Fig.~\ref{fig:model}, begins with a hyperspectral image $\mathbf{I}_{\text{Original}} \in \mathbb{R}^{H \times W \times C}$. To reduce spectral redundancy and computational cost, PCA is applied, yielding a compressed image $\mathbf{I}_{\text{Reduced}} \in \mathbb{R}^{H \times W \times P}$, where $P \ll C$. Patches of size $M \times M \times P$ are extracted and passed through two 3D convolutional layers with ReLU activations to jointly capture local spectral and spatial features.

The resulting features are processed by two parallel MLP-style mixer modules. The spectral mixer block captures long-range dependencies across spectral bands using fully connected layers and residual connections, while the spatial mixer block models spatial interactions through a similar structure applied across spatial positions. Their outputs are concatenated to form a unified spectral-spatial representation.

A lightweight attention mechanism based on depthwise convolution and sigmoid activation is then applied to generate a channel-wise attention mask, which modulates the feature maps through element-wise multiplication. Finally, global average pooling is applied, and the resulting features are fed into a softmax classifier to predict the class labels.

 \vspace{-1em}

\subsection{Feature Extraction using 3D-CNN}
To jointly capture spatial and spectral correlations in hyperspectral data, two 3D convolutional layers with $3 \times 3 \times 3$ kernels and ReLU activation are employed at the network’s input stage. Unlike 2D-CNNs, 3D-CNNs process height, width, and spectral dimensions simultaneously, enabling the extraction of local spatial structures while preserving spectral continuity. This results in rich low-level feature embeddings that support subsequent spectral-spatial mixing and attention mechanisms.

\subsection{Spectral Mixer Block}
The spectral mixer block is designed to capture long-range dependencies across spectral bands by treating each band as a token and applying multilayer perceptrons (MLPs) to mix information along the spectral dimension. As illustrated in Fig.~\ref{fig:model}(b), the input tensor of shape $M \times M \times P \times D$ is first reshaped into $(M \cdot M) \times D \times P$, aligning the spectral bands across the last axis for each spatial location.

For each of the $M \cdot M$ spatial positions, a two-layer MLP is applied independently across the $P$ spectral components to learn inter-band relationships:

\begin{equation}
\label{eq:spectral_mlp}
\mathbf{Z} = \mathbf{X} + \psi_2\left(\psi_1(\mathbf{X})\right),
\end{equation}

\noindent where $\mathbf{X} \in \mathbb{R}^{(M \cdot M) \times D \times P}$, and $\psi_1: \mathbb{R}^{P} \rightarrow \mathbb{R}^{h}$, $\psi_2: \mathbb{R}^{h} \rightarrow \mathbb{R}^{P}$ are fully connected layers operating along the spectral axis, with $h$ denoting the hidden dimension (e.g., $h = 128$). The residual connection preserves the original spectral information while allowing for deeper representations.

This spectral mixing operation is repeated over multiple blocks (e.g., four), and the output is reshaped back to $M \times M \times P \times D$ for downstream processing. The block thus enables effective learning of non-local spectral features without relying on convolutional or attention-based mechanisms.

\subsection{Spatial Mixer Block}
\vspace{-0.5em}
The spatial mixer block aims to model long-range spatial dependencies by treating each spatial location within a patch as a token and mixing spatial information across all positions using MLP layers. As illustrated in Fig.~\ref{fig:model}(c), the input feature tensor has a shape of $M \times M \times P \times D$, where $M \times M$ denotes the spatial patch size, $P$ is the reduced spectral dimension, and $D$ is the number of feature channels (32 in this case).

To prepare the data for spatial mixing, the tensor is first permuted and reshaped to a new shape of $P \times D \times (M \cdot M)$, effectively organizing spatial tokens across the last dimension. For each of the $P \times D$ tokens, a two-layer MLP is applied to mix information across the $M \times M$ spatial positions:

\begin{equation}
\label{eq:spatial_mlp}
\mathbf{Z} = \mathbf{X} + \phi_2\left(\phi_1(\mathbf{X})\right),
\end{equation}

\noindent where $\mathbf{X} \in \mathbb{R}^{P \times D \times (M \cdot M)}$, $\phi_1: \mathbb{R}^{M^2} \rightarrow \mathbb{R}^{h}$ and $\phi_2: \mathbb{R}^{h} \rightarrow \mathbb{R}^{M^2}$ are fully connected layers with hidden dimension $h$ (set to 128 in this case), and the residual connection preserves the original signal.

This MLP-based mixing is repeated $L$ times (here, $L=4$) to deepen the model's capacity for learning spatial relationships. After the mixing process, the tensor is reshaped back to its original spatial format $M \times M \times P \times D$ for downstream processing. By leveraging global spatial interactions through MLPs, this block enhances the network’s ability to capture non-local dependencies without the need for attention or convolutional operations.

\subsection{Channel-wise Spatial Attention}
\begin{figure}[!t]
\centering
\includegraphics[clip=true, trim = 40 10 40 5,width= 0.9\linewidth]{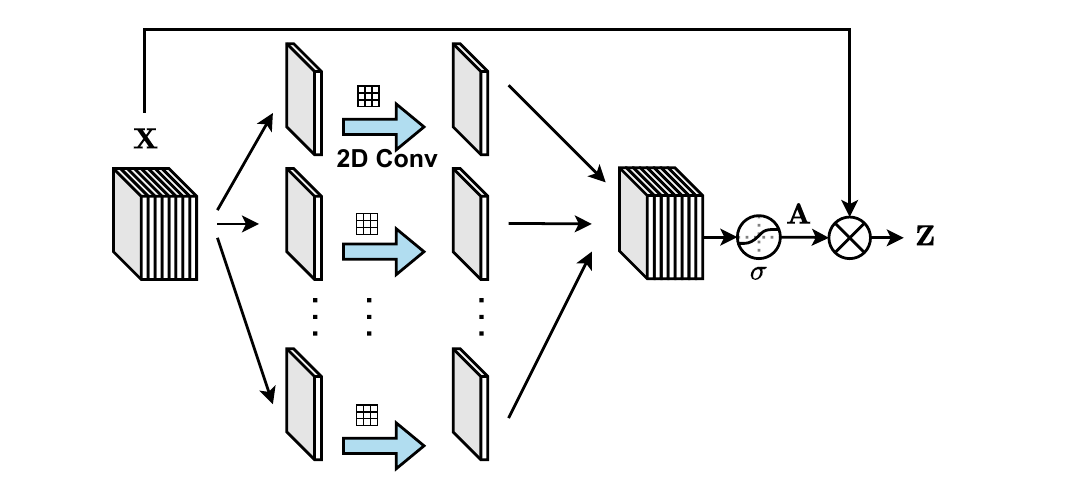}
\caption{Depthwise Attention Block}
\vspace{-1em}
 \label{fig:dwConv}
\end{figure}

Depthwise convolution functions as a lightweight attention mechanism by processing each input channel independently to learn spatial importance without inter-channel mixing. Each channel is convolved with its own spatial kernel, preserving the number of channels while capturing spatial dependencies efficiently. A subsequent sigmoid activation generates an attention mask that modulates the input feature map through element-wise multiplication. This process enables the network to emphasize informative spatial regions with minimal computational cost, as illustrated in Fig.~\ref{fig:dwConv}.

\section{Experiments and Analysis} 
\label{sec:results}
\begin{figure} [t]
   \centering
   \includegraphics[width= 0.85\linewidth]{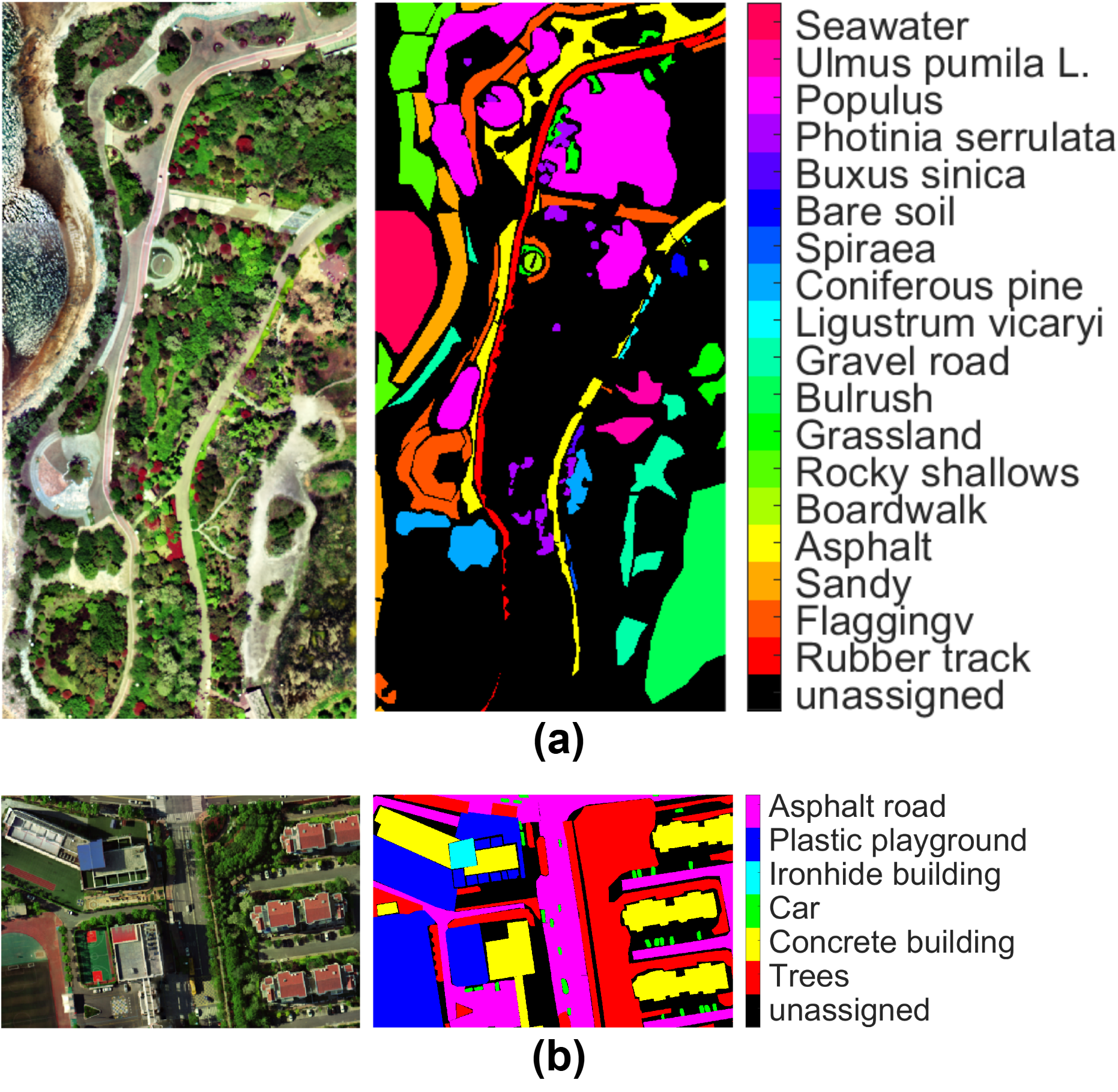}
   \vspace{-1em}
   \caption{ \label{fig:datasets}Reference Data: (a) QUH-Tangdaowan; (b) QUH-Qingyun.}
   \end{figure}

\vspace{-0.75em}
To evaluate the performance of the proposed model (Fig.~\ref{fig:model}), comparisons are made with 2D-CNN \cite{makantasis2015deep}, 3D-CNN \cite{hamida20183}, IP SWIN \cite{liu2023spectral}, SimPoolFormer \cite{roy2025simpoolformer}, and HybridKAN \cite{jamali2024learn}. The evaluation metrics include Overall Accuracy (OA), Average Accuracy (AA), Kappa coefficient, and per-class accuracy. Experiments are conducted on two widely used hyperspectral datasets: QUH-Tangdaowan and QUH-Qingyun. The corresponding reference maps are shown in Fig.~\ref{fig:datasets}, with dataset details provided in \cite{jamali2024learn}.

For both datasets, patches were randomly split into 1\% training, 1\% validation, and 98\% testing. A patch size of $9 \times 9$ and 15 principal components were used. The model was trained for 100 epochs with a batch size of 64 using the Adam optimizer (learning rate $1 \times 10^{-3}$). Early stopping was applied, halting training if no improvement was observed over 10 consecutive epochs and restoring the best weights. All models were implemented in Python using Keras with TensorFlow 2.10.0, and trained under identical settings to ensure fair comparison.

\begin{figure} [t!]
   \centering
   \includegraphics[width= 0.85\linewidth]{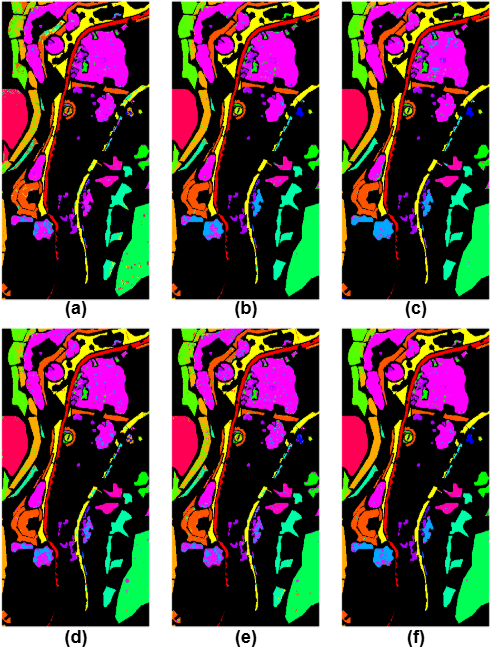}  
   \vspace{-1em}
   \caption{ \label{fig:tngdn_results} Classification maps of Tangdaowan Dataset. (a) 2D-CNN; (b) 3D-CNN; (c) IP-SWIN; (d) SimPoolFormer (e) HybridKAN; (f) Proposed.}
   \end{figure}

Table~\ref{tab:tngdn} presents the classification performance of various models on the Tangdaowan dataset. The proposed SS-MixNet achieves the highest OA (95.68\%), AA (92.44\%), and Kappa (95.08\%), outperforming all compared methods, including 3D-CNN (94.50\% OA, 90.11\% AA) and IP-SWIN (94.32\% OA, 88.36\% AA). Notably, SS-MixNet achieves the best classification accuracy in 11 out of the 18 classes, demonstrating its effectiveness in capturing both spectral and spatial features. It performs particularly well in challenging categories such as \textit{Sandy}, \textit{Asphalt}, \textit{Rocky shallows}, and \textit{Populus}. While other models show strong performance in specific classes—for example, 3D-CNN on \textit{Boardwalk} and \textit{Bare soil}—SS-MixNet provides more consistent and superior results across the majority of classes. Transformer-based classifiers, such as SimPoolFormer, showed lower performance, likely due to their dependence on large amounts of training data, which is limited in this setting. The corresponding classification maps for the Tangdaowan dataset are shown in Fig.~\ref{fig:tngdn_results}, further illustrating the visual quality of the predicted outputs.

\begin{table}[t!]
\centering
\caption{Classification performance of different methods for the Tangdaowan dataset. Bold indicates the best result}
\vspace{-1em}
\label{tab:tngdn}
\resizebox{\linewidth}{!}{
\begin{tabular}{cccccccccc}
\hline
\textbf{Class}     & \textbf{Train} & \textbf{Val} & \textbf{Test} & \textbf{\begin{tabular}[c]{@{}c@{}}2D-\\ CNN\end{tabular}} & \textbf{\begin{tabular}[c]{@{}c@{}}3D-\\ CNN\end{tabular}} & \textbf{\begin{tabular}[c]{@{}c@{}}IP-\\ SWIN\end{tabular}} & \textbf{\begin{tabular}[c]{@{}c@{}}SimPool\\ Former\end{tabular}} & \textbf{\begin{tabular}[c]{@{}c@{}}Hybrid\\ KAN\end{tabular}} & \textbf{\begin{tabular}[c]{@{}c@{}}SS-Mix\\ Net\end{tabular}} \\ \hline
Rubber track       & 258            & 258          & 25,333        & 98.74                                                      & 98.98                                                      & \textbf{99.83}                                              & 99.61                                                             & 98.38                                                         & 99.78                                                         \\
Flaggingv          & 555            & 555          & 54,443        & 90.25                                                      & \textbf{98.85}                                             & 98.34                                                       & 97.16                                                             & 92.91                                                         & 98.45                                                         \\
Sandy              & 340            & 340          & 33,357        & 87.11                                                      & 92.45                                                      & 92.66                                                       & 92.84                                                             & 78.55                                                         & \textbf{95.08}                                                \\
Asphalt            & 607            & 607          & 59,476        & 94.61                                                      & 98.35                                                      & 98.61                                                       & 92.73                                                             & 93.68                                                         & \textbf{99.18}                                                \\
Boardwalk          & 19             & 19           & 1,824         & 35.18                                                      & \textbf{89.80}                                             & 83.46                                                       & 87.54                                                             & 71.00                                                         & 86.31                                                         \\
Rocky shallows     & 371            & 371          & 36,383        & 64.78                                                      & 90.07                                                      & 88.16                                                       & 83.53                                                             & 87.07                                                         & \textbf{91.40}                                                \\
Grassland          & 141            & 141          & 13,845        & 63.35                                                      & 77.38                                                      & 77.39                                                       & 72.45                                                             & 62.21                                                         & \textbf{77.94}                                                \\
Bulrush            & 641            & 641          & 62,805        & 96.38                                                      & 99.86                                                      & 99.48                                                       & 98.40                                                             & 98.57                                                         & \textbf{99.80}                                                \\
Gravel road        & 307            & 307          & 30,081        & 90.11                                                      & 97.54                                                      & 98.34                                                       & 97.81                                                             & 93.85                                                         & \textbf{98.54}                                                \\
Ligustrum vicaryi  & 18             & 18           & 1,747         & 55.36                                                      & 95.96                                                      & \textbf{98.93}                                              & 76.28                                                             & 80.15                                                         & 96.52                                                         \\
Coniferous pine    & 212            & 212          & 20,812        & 26.80                                                      & 74.88                                                      & \textbf{88.93}                                              & 43.64                                                             & 34.86                                                         & 81.35                                                         \\
Spiraea            & 8              & 8            & 733           & 67.16                                                      & 68.89                                                      & 18.83                                                       & 73.30                                                             & 50.87                                                         & \textbf{80.64}                                                \\
Bare soil          & 17             & 17           & 1,652         & 47.86                                                      & \textbf{99.94}                                             & 99.88                                                       & 41.28                                                             & 86.36                                                         & 99.64                                                         \\
Buxus sinica       & 9              & 9            & 868           & 69.19                                                      & 82.28                                                      & 86.91                                                       & 72.23                                                             & 42.44                                                         & \textbf{93.91}                                                \\
Photinia serrulata & 140            & 140          & 13,740        & 76.78                                                      & \textbf{85.66}                                             & 83.79                                                       & 68.98                                                             & 64.38                                                         & 83.22                                                         \\
Populus            & 1,409          & 1,409        & 138,086       & 92.97                                                      & 94.18                                                      & 91.70                                                       & 92.13                                                             & 88.35                                                         & \textbf{95.80}                                                \\
Ulmus pumila L     & 98             & 98           & 9,606         & 62.08                                                      & 76.97                                                      & 85.59                                                       & 76.76                                                             & 64.09                                                         & \textbf{86.30}                                                \\
Seawater           & 423            & 423          & 41,429        & 97.63                                                      & 99.93                                                      & 99.64                                                       & 99.60                                                             & 99.38                                                         & \textbf{99.98}                                                \\ \hline
\multicolumn{4}{c}{OA (\%)}                                        & 86.75                                                      & 94.50                                                      & 94.32                                                       & 90.64                                                             & 87.54                                                         & \textbf{95.68}                                                \\
\multicolumn{4}{c}{AA (\%)}                                        & 73.13                                                      & 90.11                                                      & 88.36                                                       & 81.46                                                             & 77.06                                                         & \textbf{92.44}                                                \\
\multicolumn{4}{c}{Kappa ($\times 100$)}              & 84.80                                                      & 93.73                                                      & 93.54                                                       & 89.29                                                             & 85.75                                                         & \textbf{95.08}                                                \\                                 
\hline
\end{tabular}}
\end{table}

\begin{figure} [t!]
   \centering
   \includegraphics[width= 0.85\linewidth]{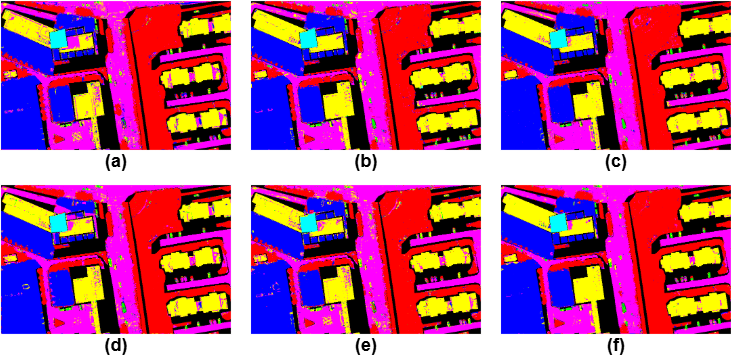}  
   \vspace{-1em}
   \caption{ \label{fig:qngn_results} Classification maps of Qingyun Dataset. (a) 2D-CNN; (b) 3D-CNN; (c) IP-SWIN; (d) SimPoolFormer (e) HybridKAN; (f) Proposed.}
   \end{figure}

   Table~\ref{tab:qngn} presents the classification results of different models on the Qingyun dataset. The proposed SS-MixNet achieves the best OA (93.86\%), AA (86.83\%), and Kappa (91.86), outperforming all competing methods. It achieves the highest class-wise accuracy in three out of six categories, including Trees, Concrete building, and Asphalt road. Notably, for dominant classes such as Trees and Concrete building, SS-MixNet reaches 95.60\% and 94.63\% accuracy, respectively. While IP-SWIN performs best in the Car class (48.97\%), and 2D-CNN slightly outperforms others in Ironhide building (98.16\%), SS-MixNet maintains consistently high performance across most categories. These results confirm the model's effectiveness in leveraging spectral and spatial features for hyperspectral image classification in complex urban environments. The corresponding classification maps for the Qingyun dataset are shown in Fig.~\ref{fig:qngn_results}, further supporting the quantitative findings through visual evidence.

\begin{table}[t!]
\centering
\caption{Classification performance of different methods for the Qingyun dataset. Bold indicates the best result}
\vspace{-1em}
\label{tab:qngn}
\resizebox{\linewidth}{!}{
\begin{tabular}{cccccccccc}
\hline
\textbf{Name}      & \textbf{Train} & \textbf{Val} & \textbf{Test} & \textbf{\begin{tabular}[c]{@{}c@{}}2D-\\ CNN\end{tabular}} & \textbf{\begin{tabular}[c]{@{}c@{}}3D-\\ CNN\end{tabular}} & \textbf{\begin{tabular}[c]{@{}c@{}}IP-\\ SWIN\end{tabular}} & \textbf{\begin{tabular}[c]{@{}c@{}}SimPool\\ Former\end{tabular}} & \textbf{\begin{tabular}[c]{@{}c@{}}Hybrid\\ KAN\end{tabular}} & \textbf{\begin{tabular}[c]{@{}c@{}}SS-Mix\\ Net\end{tabular}} \\ \hline
Trees              & 2,781          & 2,781        & 272,588       & 94.95                                                      & 93.81                                                      & 93.32                                                       & 95.08                                                             & 93.99                                                         & \textbf{95.60}                                                \\
Concrete building  & 1,795          & 1,795        & 175,922       & 81.54                                                      & 93.95                                                      & 94.18                                                       & 88.47                                                             & 90.08                                                         & \textbf{94.63}                                                \\
Car                & 138            & 138          & 13,507        & 26.87                                                      & 43.29                                                      & \textbf{48.97}                                              & 42.32                                                             & 21.72                                                         & 45.00                                                         \\
Ironhide building  & 98             & 98           & 9,571         & \textbf{98.16}                                             & 96.81                                                      & 97.54                                                       & 97.43                                                             & 91.71                                                         & 97.75                                                         \\
Plastic playground & 2,177          & 2,177        & 213,381       & 92.30                                                      & 95.07                                                      & \textbf{95.98}                                              & 89.31                                                             & 91.38                                                         & 95.46                                                         \\
Asphalt road       & 2,559          & 2,559        & 250,828       & 85.41                                                      & 90.04                                                      & 92.00                                                       & 90.56                                                             & 87.00                                                         & \textbf{92.55}                                                \\ \hline
\multicolumn{4}{c}{OA (\%)}                                        & 88.32                                                      & 92.42                                                      & 93.14                                                       & 90.57                                                             & 89.72                                                         & \textbf{93.86}                                                \\
\multicolumn{4}{c}{AA (\%)}                                        & 79.87                                                      & 85.50                                                      & 86.55                                                       & 83.86                                                             & 79.31                                                         & \textbf{86.83}                                                \\
\multicolumn{4}{c}{Kappa ($\times 100$}                            & 84.48                                                      & 89.96                                                      & 90.90                                                       & 87.48                                                             & 86.36                                                         & \textbf{91.86}                                                \\ \hline
\end{tabular}
}
 \vspace{-1em}
\end{table}

The ablation study in Table~\ref{tab:ablation} demonstrates the incremental contribution of each module to the overall accuracy (OA) on the Tangdaowan dataset. Starting from a baseline 3D-CNN with 94.20\% OA, the inclusion of either the spectral or spatial mixer block individually improves performance to 95.07\% and 94.89\%, respectively. When both mixers are combined, OA increases to 95.38\%, confirming their complementary effects. Adding the attention mechanism further boosts OA to 95.68\%, highlighting its role in enhancing feature discrimination and confirming the effectiveness of the complete SS-MixNet architecture.

\begin{table}[t!]
\centering
\caption{Results of ablation studies on different combinations of model
components applied to the Tangdaowan dataset.}
\vspace{-1em}
\label{tab:ablation}
\resizebox{\linewidth}{!}{
\begin{tabular}{lccc}
\hline
Combination                    & OA (\%) & AA (\%) & Kappa $\times 100$ \\ \hline
3D-CNN                         & 94.20   & 89.81   & 93.43              \\
3D-CNN + (Spe)ctral            & 95.07   & 90.67   & 94.38              \\
3D-CNN + (Spa)tial             & 94.89   & 90.53   & 94.18              \\
3D-CNN + Spe + Spa             & 95.38   & 92.37   & 94.74              \\
3D-CNN + Spe + Spa + Attention & 95.68   & 92.44   & 95.08              \\ \hline
\end{tabular}
}
 \vspace{-1.5em}
\end{table}

Table~\ref{tab:efficiency} summarizes the computational efficiency of all models considered in this study in terms of the number of parameters, floating-point operations (FLOPs), and multiply–accumulate operations (MACs). Among the compared architectures, SimPoolFormer exhibits the highest computational cost, requiring over 57 million FLOPs and 28 million MACs due to its complex pooling and attention mechanisms. Conversely, IP-SWIN has the smallest parameter count (approximately 90 K), indicating a lightweight design despite its moderate computational demand. The proposed SS-MixNet achieves a balanced trade-off, with 141 K parameters and 1.9 M FLOPs, offering efficiency while maintaining strong representation capability. The 2D-CNN and 3D-CNN models demonstrate relatively low complexity, whereas HybridKAN presents higher computational requirements but improved expressive power. Overall, the table highlights that SS-MixNet provides an effective balance between computational efficiency and modeling capacity compared to both conventional CNNs and transformer-based counterparts.

\begin{table}[tb!]
\centering
\caption{Parameters, FLOPs, and MACs of each Model used in the research}
\vspace{-1em}
\label{tab:efficiency}
\resizebox{0.9\linewidth}{!}{
\begin{tabular}{lccc}
\hline
Model         & Parameters & FLOPs      & MACs       \\ \hline
2D-CNN        & 287,950    & 1,658,448  & 829,224    \\
3D-CNN        & 397,586    & 682,240    & 341,120    \\
IP-SWIN       & 90,154     & 1,985,489  & 988,288    \\
SimPoolFormer & 771,122    & 57,497,293 & 28,423,255 \\
HybridKAN     & 142,690    & 20,200,395 & 10,027,719 \\
SS-MixNet     & 140,914    & 1,932,831  & 771,408    \\ \hline
\end{tabular}}
\end{table}

\section{Conclusion}\label{sec:con}
\vspace{-0.75em}
In this paper, a novel and lightweight architecture, SS-MixNet, was proposed for hyperspectral image classification. The model effectively combines 3D convolutional layers with spectral and spatial MLP-style mixer blocks to jointly capture local and long-range dependencies in both spectral and spatial dimensions. Additionally, a depthwise convolution-based attention mechanism was introduced to enhance feature representation with minimal computational overhead. The proposed model was evaluated on two challenging hyperspectral datasets, QUH-Tangdaowan and QUH-Qingyun, using a limited supervision setting with only 1\% of labeled data for training and validation. SS-MixNet consistently outperformed several state-of-the-art methods in terms of overall accuracy, average accuracy, and Kappa coefficient, achieving 95.68\% and 93.86\% OA on the respective datasets.

These results confirm the model's effectiveness in learning discriminative and robust spectral–spatial representations, even under conditions of severe label scarcity. In future work, the framework will be extended to address cross-domain HSI classification and evaluated under semi-supervised and self-supervised learning settings. Furthermore, the integration of hybrid designs incorporating transformer-based modules with learnable prompts will be explored to enhance adaptability and generalization across diverse remote sensing scenarios.

\patchcmd{\thebibliography}{\settowidth}{\setlength{\itemsep}{0pt}\setlength{\parskip}{0pt plus 0.0ex}\settowidth}{}{}
\bibliographystyle{IEEEbib}
\bibliography{Main_Doc}

\end{document}